\documentclass[twocolumn,compsoc,journal]{IEEEtran}
\usepackage[T1]{fontenc}
\usepackage[latin9]{inputenc}
\usepackage{float}
\usepackage{textcomp}
\usepackage{mathrsfs}
\usepackage{amsmath}
\usepackage{amssymb}
\usepackage{graphicx}
\usepackage[unicode=true,
 bookmarks=true,bookmarksnumbered=true,bookmarksopen=true,bookmarksopenlevel=1,
 breaklinks=false,pdfborder={0 0 0},pdfborderstyle={},backref=false,colorlinks=false]
 {hyperref}
\hypersetup{pdftitle={Your Title},
 pdfauthor={Your Name},
 pdfpagelayout=OneColumn, pdfnewwindow=true, pdfstartview=XYZ, plainpages=false}

\makeatletter

\providecommand{\tabularnewline}{\\}

\usepackage[caption=false,font=footnotesize,labelfont=sf,textfont=sf]{subfig}
\usepackage[mathscr]{euscript}

\@ifundefined{showcaptionsetup}{}{%
 \PassOptionsToPackage{caption=false}{subfig}}
\usepackage{subfig}
\makeatother

\begin{document}
\title{PrognoseNet: A Generative Probabilistic Framework for Multimodal Position
Prediction given Context Information}
\author{Thomas~Kurbiel,~Akash\ Sachdeva,\ Kun\ Zhao\ and\ Markus\ Buehren\\
$\phantom{}$\\
\textit{Aptiv GmbH, Wuppertal, Germany}\\
\IEEEcompsocitemizethanks{\IEEEcompsocthanksitem e-mail: \href{http://Thomas.Kurbiel\%40aptiv.com}{Thomas.Kurbiel@aptiv.com},
\href{http://Akash.Sachdeva\%40aptiv.com}{Akash.Sachdeva@aptiv.com}\protect 

\ \ \ \ \ \ \ \ \ \ \ \href{http://Kun.Zhao\%40aptiv.com}{Kun.Zhao@aptiv.com},
\href{http://Markus.Buehren\%40aptiv.com}{Markus.Buehren@aptiv.com}\protect \\

}}

\IEEEtitleabstractindextext{
\begin{abstract}
The ability to predict multiple possible future positions of the ego-vehicle
given the surrounding context while also estimating their probabilities
is key to safe autonomous driving. Most of the current state-of-the-art
Deep Learning approaches are trained on trajectory data to achieve
this task. However trajectory data captured by sensor systems is highly
imbalanced, since by far most of the trajectories follow straight
lines with an approximately constant velocity. This poses a huge challenge
for the task of predicting future positions, which is inherently a
regression problem. Current state-of-the-art approaches alleviate
this problem only by major preprocessing of the training data, e.g.
resampling, clustering into anchors etc.

In this paper we propose an approach which reformulates the prediction
problem as a classification task, allowing for powerful tools, e.g.
focal loss, to combat the imbalance. To this end we design a generative
probabilistic model consisting of a deep neural network with a Mixture
of Gaussian head. A smart choice of the latent variable allows for
the reformulation of the log-likelihood function as a combination
of a classification problem and a much simplified regression problem.
The output of our model is an estimate of the probability density
function of future positions, hence allowing for prediction of multiple
possible positions while also estimating their probabilities. The
proposed approach can easily incorporate context information and does
not require any preprocessing of the data.
\end{abstract}


}
\maketitle

\section{Introduction\label{sec:Introduction}}

Human drivers possess the fundamental skill of predicting a multitude
of different possible future positions and movements of other traffic
participants. Their predictions highly depend on the surroundings
and interactions between traffic participants. This human ability
ensures safe and efficient driving. Reproducing this ability by machines
is key to safe autonomous driving. Due to the highly dynamic and complex
driving environment a large amount of training data is needed in order
to develop a system capable of operating at a level comparable to
human drivers.

\subsubsection*{Trajectory data}

Most of the current state-of-the-art Deep Learning approaches are
trained using trajectory data: spatio-temporal data capturing the
movement of vehicles, pedestrians etc. over a certain temporal interval.
Trajectory data is usually organized into sequences with three or
more dimensions, e.g. batch, time, features. The features can consist
of ego x- and y-coordinates, heading angles, velocities etc. To reflect
the interaction with other agents, trajectory data may contain multiple
objects. Furthermore to reflect the interaction of the objects with
the environment, a static map may be included. Trajectory data can
be synthetic or based on a real data set.

An eligible approach has to cope with the following two challenges
posed by trajectory data captured by sensor systems.

\subsubsection*{Imbalanced dataset}

\begin{figure}[H]
\begin{centering}
\subfloat[ego-position in 2000ms]{\includegraphics[viewport=5bp 0bp 267bp 260bp,scale=0.48]{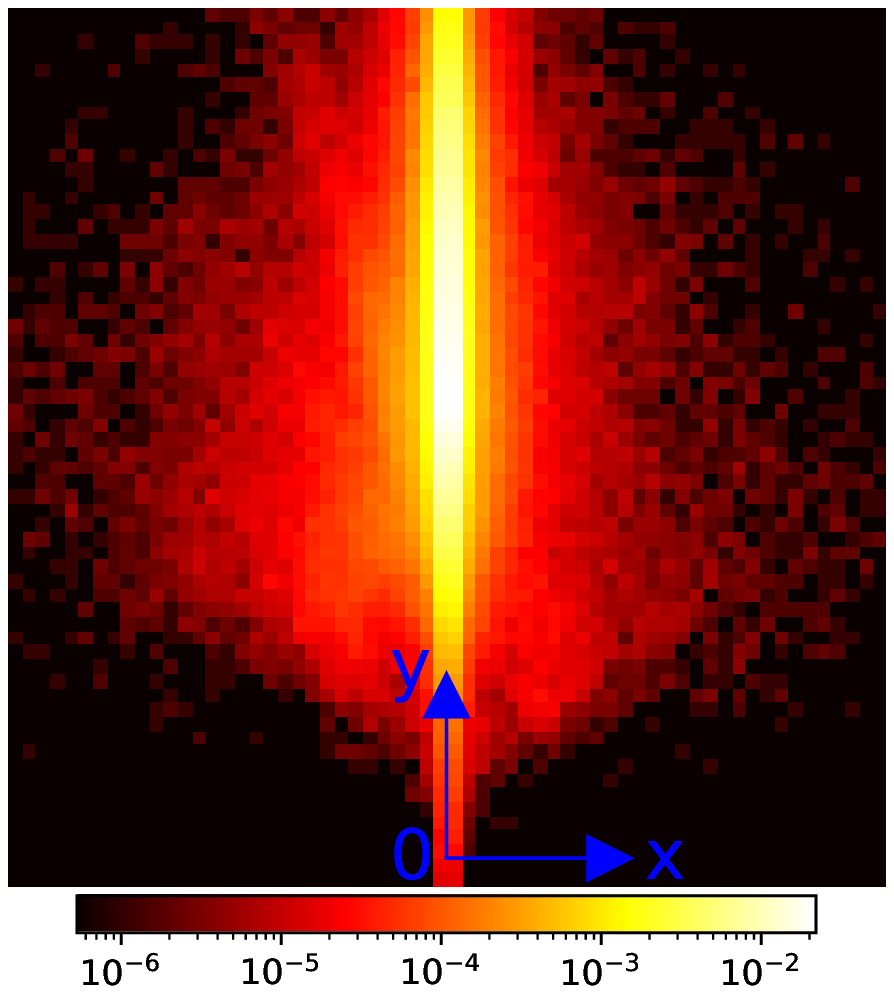}

}\subfloat[ego-position in 4000ms]{\includegraphics[viewport=10bp 0bp 267bp 260bp,scale=0.48]{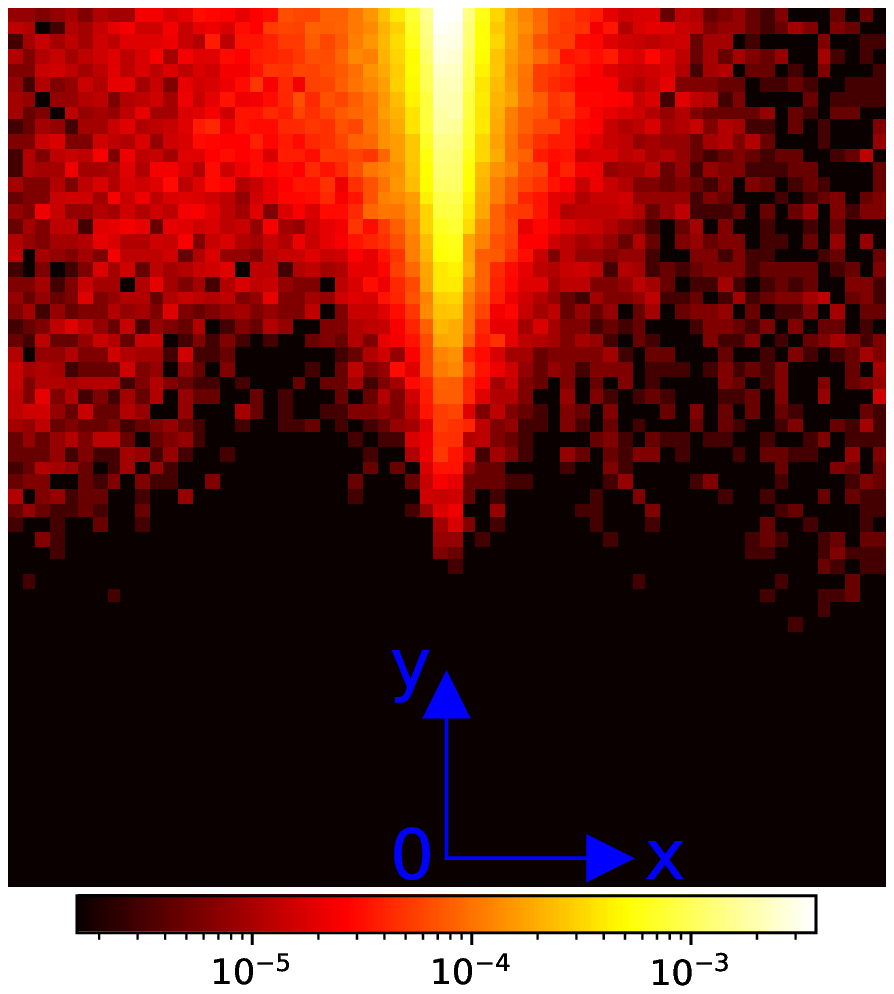}

}
\par\end{centering}
\caption{Illustration of the dominance of straight lines trajectories\newline$\protect\phantom{ddddd}$The
ego-vehicle at current time step is at the origin of the VCS\label{fig:imbalanced-dataset}}
\end{figure}

Trajectory data captured by sensor systems is highly imbalanced, since
by far most of the trajectories follow straight lines with an approximately
constant velocity. However it is the abnormal behaviors: \textquotedblleft unexpected
stops\textquotedblright , \textquotedblleft accelerations\textquotedblright ,
\textquotedblleft turnings\textquotedblright , \textquotedblleft deviation
from standard routes\textquotedblright{} which interest us and pose
a challenge. Fig.\ \ref{fig:imbalanced-dataset} depicts the probabilities
of future ego-positions evaluated on the Argoverse dataset \cite{Chang2019}.
The evaluation was performed in the vehicle coordinate system (VCS).

Fig.\ \ref{fig:imbalanced-dataset} clearly shows how straight line
trajectories dominate the dataset. Please note that the colorbar is
scaled logarithmically in order to make future ego-positions (not
lying on the straight line) visible in the first place. Their probability
is several orders of magnitude lower than that of the straight line
trajectories.

\subsubsection*{Multimodality}

The second challenge consists in predicting not only the most probable
future position, but a multitude of different possible future positions
and assigning a probability to each one of them. The multimodality
of future positions occurs due to many factors, e.g. multiple possible
paths, different acceleration patterns, interactions with other traffic
participants, just to name a few.

\section{Related Work\label{sec:Related-Work}}

In the past, many sophisticated methods have been developed in the
field of trajectory prediction. Due to the ongoing research, new approaches
are presented almost on a weekly basis. In this section we will therefore
only mention the most important ones and illustrate how good or bad
they deal with the problems described in the previous section.

One means to cope with imbalanced data is to resample the data, e.g.
remove some observations of the majority class (undersampling) or
add more copies of the minority class (oversampling) \cite{Chawla2010}.
However, the definition of the minority class and the majority class
is based on criteria, which oftentimes are hand-engineered. A drawback
of this approach is that we might remove information that is be valuable.
This could lead to underfitting and poor generalization to the test
set. Think of the example where we want to balance out the dataset
by removing trajectories which follow straight lines. We could classify
the trajectories according to their curvature, this way neglecting
the acceleration behavior and hence dropping important samples. This
simple example illustrates how difficult the definition of a proper
criterion can get. Another fact to emphasize is that even trajectories
which are characterized as curves partially consists of segments which
follow straight lines.
\begin{figure}[tbh]
\begin{centering}
\includegraphics[viewport=0bp 30bp 370bp 330bp,scale=0.35]{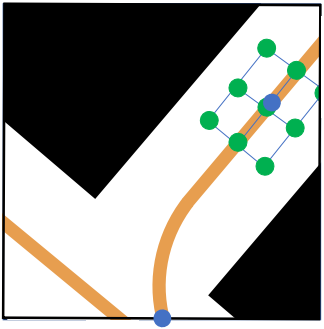}
\par\end{centering}
\caption{Predictions (green) of the min-of-$N$ lying on a uniform grid.\newline$\protect\phantom{dddddd}$Number
of latent noises is 9.\label{fig:min_of_N}}
\end{figure}

Similar arguments apply to approaches which partition the trajectories
into several clusters/anchors and treat the prediction problem as
a classification of the correct cluster/anchor \cite{Chai2019,PhanMinh}.
Here again, the clustering itself is oftentimes based on hand-engineered
criteria. Furthermore, the large number of situations encountered
on roads and the high uncertainty of traffic behavior requires a large
number of clusters. The drawback of cluster based approaches is that
their prediction of multimodality is limited due to limited number
of clusters.

The basic min-of-$N$ approach is formulated as a regression problem,
where multimodality is achieved by introducing an additional latent
noise as input \cite{Fan2016,Bhattacharyya2018}. Each choice of the
latent noise produces a different prediction. During training, optimization
is performed only on the latent noise corresponding to the prediction
with the minimal error to the ground truth position. This approach
has the drawback that the predictions tend to lie on a uniform grid
as depicted in Fig.\ \ref{fig:min_of_N} especially if the number
of latent noises is high. This way the neural network ensures, that
there is at least one prediction which lies close to the ground truth
position. The major drawback of this approach are the missing probabilities
assigned to the predictions. Besides, this approach suffers from the
imbalance of data and hence requires preprocessing of the data. The
training is computationally very expensive due to the computation
of different latent noises.

Another approach is based on Conditional Variational Autoencoders
(CVAE), which models multimodality by sampling from a conditional
Gaussian Distribution \cite{Gupta,Lee2017}. Since CVAEs do not provide
any means to inherently emphasize underrated samples, this approach
too suffers from imbalance of data. Here, the conditional probability
realized by the encoder network is dominated by the majority class.
Furthermore assigning probabilities to the predictions at the output
of the decoder is difficult, even though the conditional probability
of the latent variable is known. However, the probability density
of a transformed random variable is not easily obtained \cite{Gamarnick2008}.

\section{Our Approach}

In this section we introduce a novel approach which solves the imbalanced
data problem by formulating the prediction problem as a classification
problem and utilizing focal loss, thus not requiring any preprocessing
of the data. The multimodality is achieved by introducing a generative
probabilistic model which outputs an estimate of the probability density
function of future positions. This way we create a fully generalizable
system, which is not confined by any hand-engineered preprocessing
of the data.

\begin{figure}[tbh]
\begin{centering}
\subfloat[GCS]{\includegraphics[viewport=-5bp 0bp 308bp 290bp,scale=0.39]{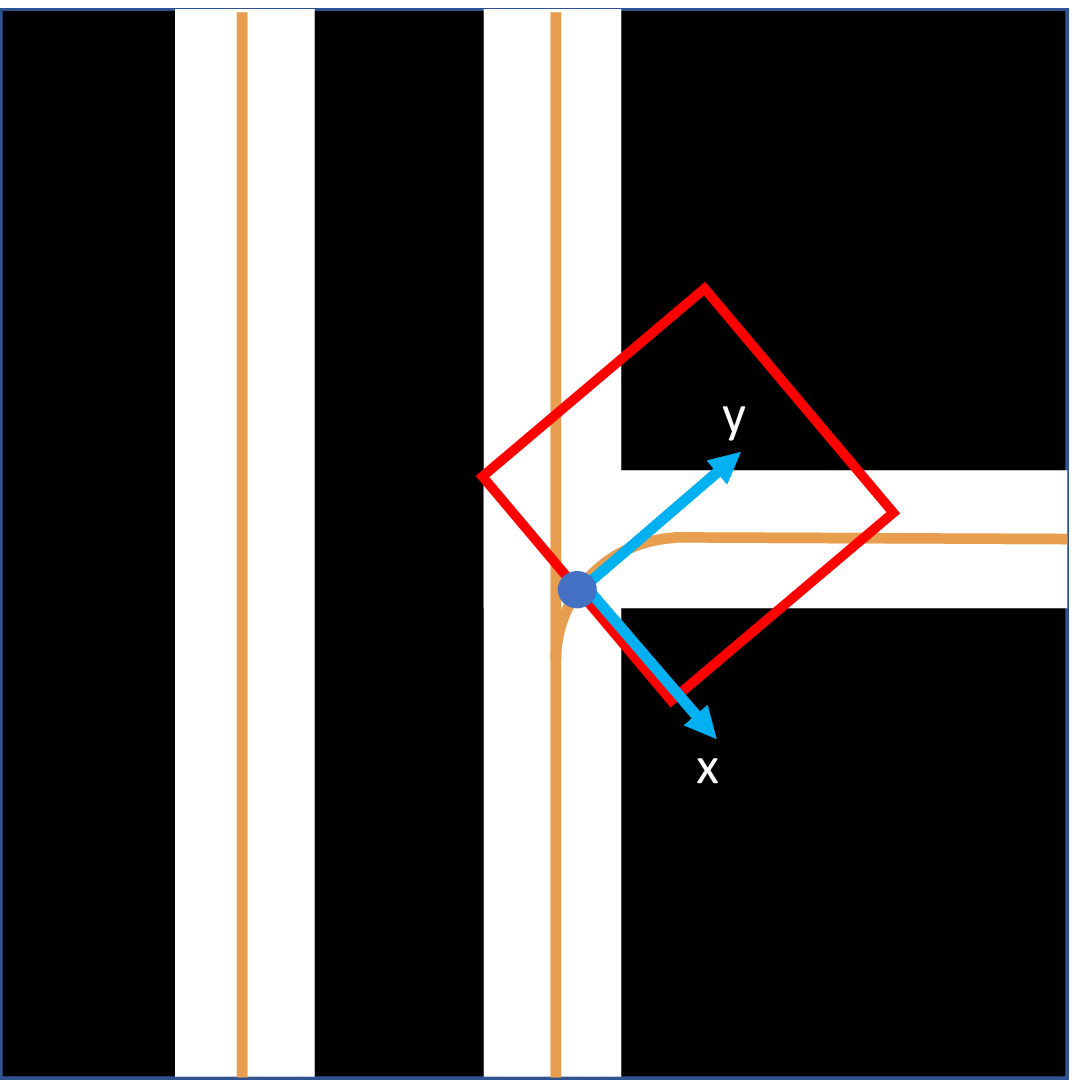}

}\subfloat[VCS]{\includegraphics[viewport=5bp 23bp 379bp 290bp,scale=0.35]{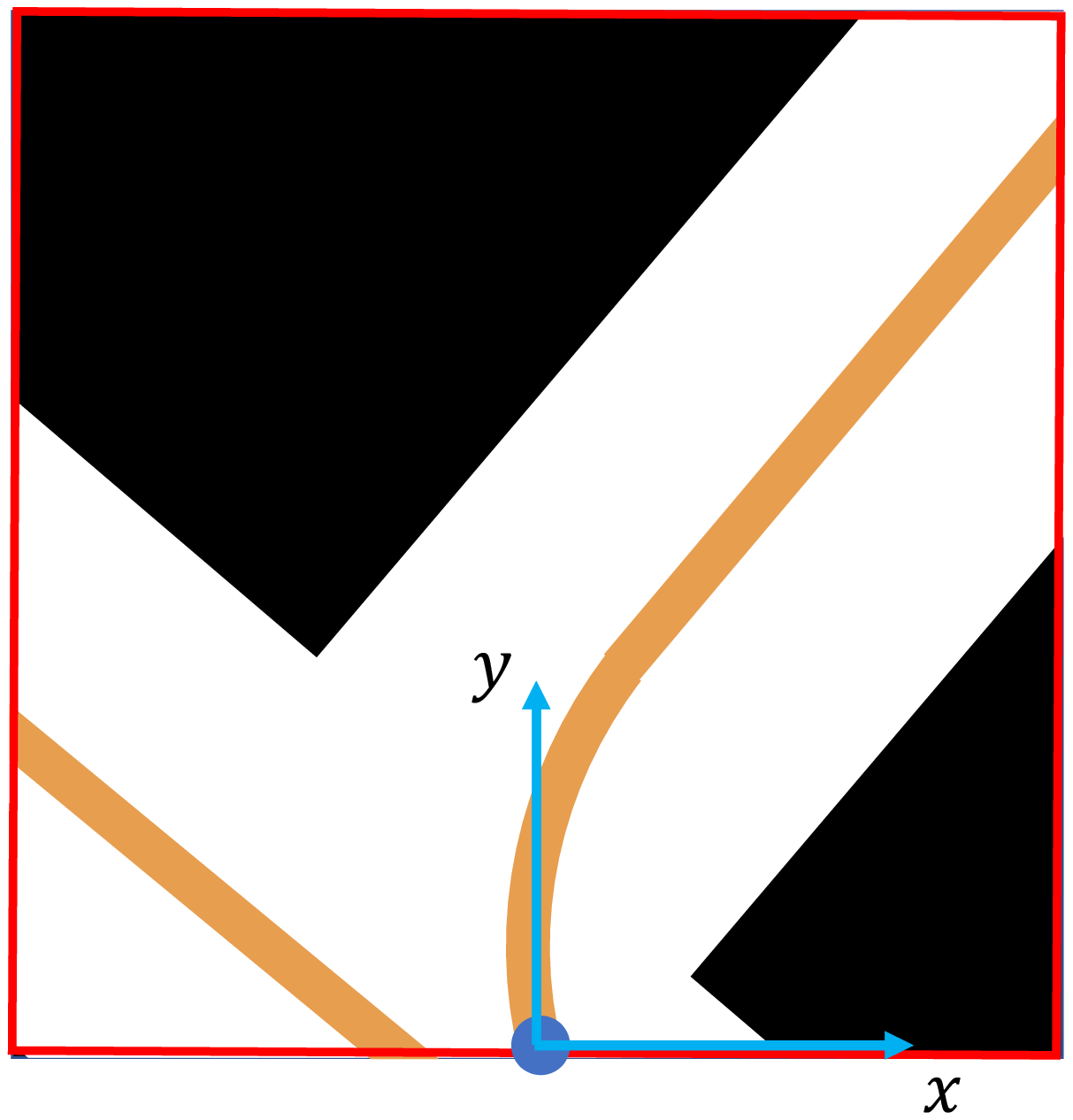}

}
\par\end{centering}
\caption{(a) Static map in global coordinate system and borders \newline$\protect\phantom{ddddddddd}$of
the static map in red \newline$\protect\phantom{dddddd}$(b) Static
map in vehicle coordinate system of time step $t$\label{fig:coordinate_systems}}
\end{figure}

\subsection{Input Output\label{sec:Input-Output}}

The input trajectory (up to time step $t$) is defined as the sequence:
\begin{equation}
\mathscr{\mathscr{T}}^{\left\langle 0:t\right\rangle }=\left\{ \left(\text{\ensuremath{\Delta}}x^{\left\langle \tau\right\rangle },\ \text{\ensuremath{\Delta}}y^{\left\langle \tau\right\rangle },\ v^{\left\langle \tau\right\rangle },\ h^{\left\langle \tau\right\rangle }\right)\right\} _{\tau=0}^{t},\label{eq:prognet_1}
\end{equation}
with $\text{\ensuremath{\Delta}}x^{\left\langle t\right\rangle }=x^{\left\langle t\right\rangle }-x^{\left\langle t-1\right\rangle }$
and $\text{\ensuremath{\Delta}}y^{\left\langle t\right\rangle }=y^{\left\langle t\right\rangle }-y^{\left\langle t-1\right\rangle }$.
For brevity in the following we will write: $\boldsymbol{x}^{\left\langle t\right\rangle }=(x^{\left\langle t\right\rangle },y^{\left\langle t\right\rangle })$.
Both $\boldsymbol{x}^{\left\langle t\right\rangle }$ and $\boldsymbol{x}^{\left\langle t-1\right\rangle }$
in the definition of the deltas are expressed in the vehicle coordinate
system of time step $t$ (\textit{please note that} $\boldsymbol{x}^{\left\langle t\right\rangle }=(0,\ 0)$
\textit{in VCS of time step} $t$). The terms $v^{\left\langle t\right\rangle }$,
$h^{\left\langle t\right\rangle }$ are the velocity and the heading
angle respectively. Please note that $t$ is an integer index denoting
measurements, which were obtained using a sampling rate of 10Hz.

The static map for time step $t$ is denoted as $\mathcal{\mathscr{M}}^{\left\langle t\right\rangle }$.
It is also expressed in the vehicle coordinate system of time step
$t$, see Fig.\ \ref{fig:coordinate_systems}. The static maps used
in this paper contain road boundaries and center lines.
\begin{figure}[tbh]
\begin{raggedleft}
\subfloat[]{\includegraphics[viewport=0bp 0bp 308bp 270bp,scale=0.39]{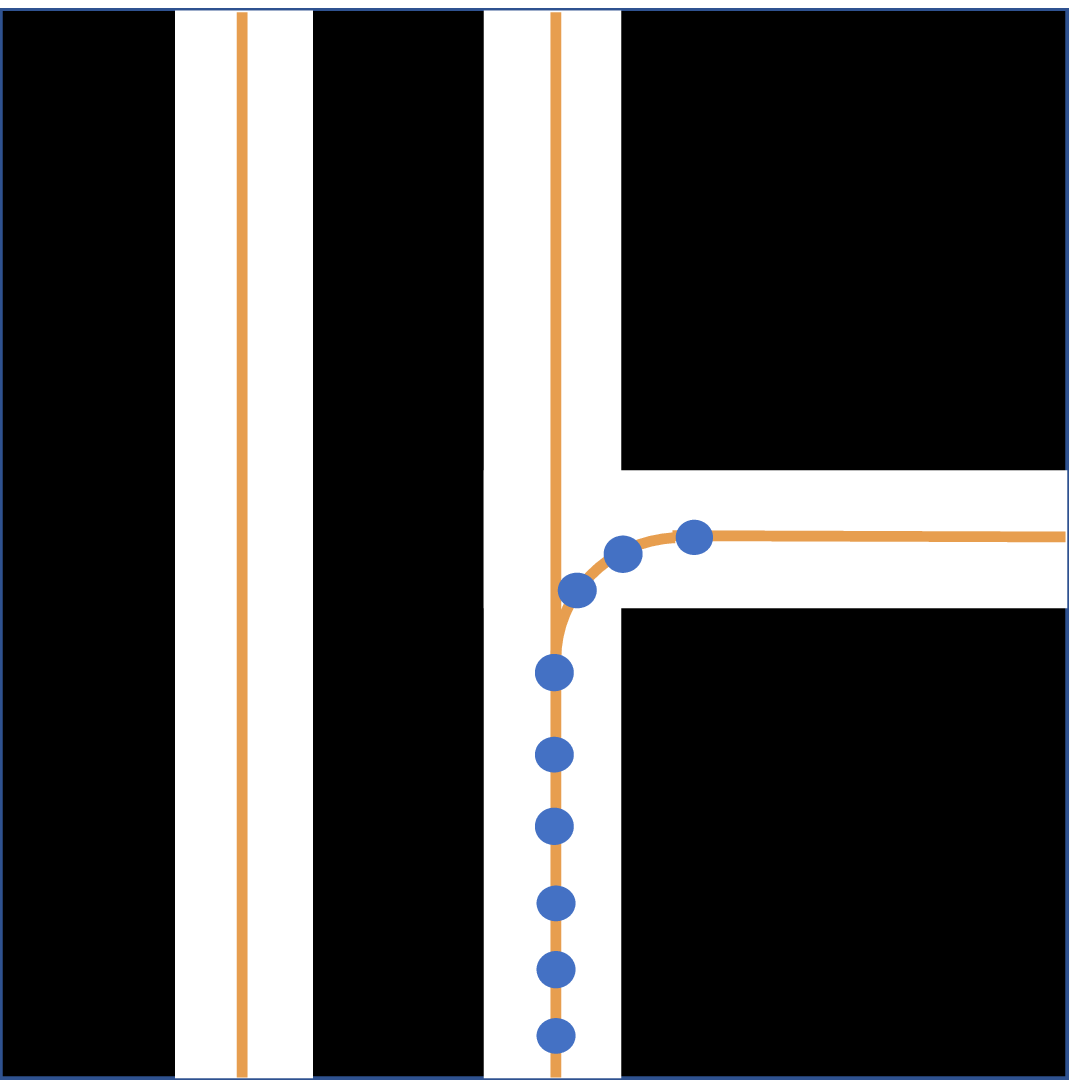}

}\subfloat[]{\includegraphics[viewport=-10bp 0bp 309bp 270bp,scale=0.39]{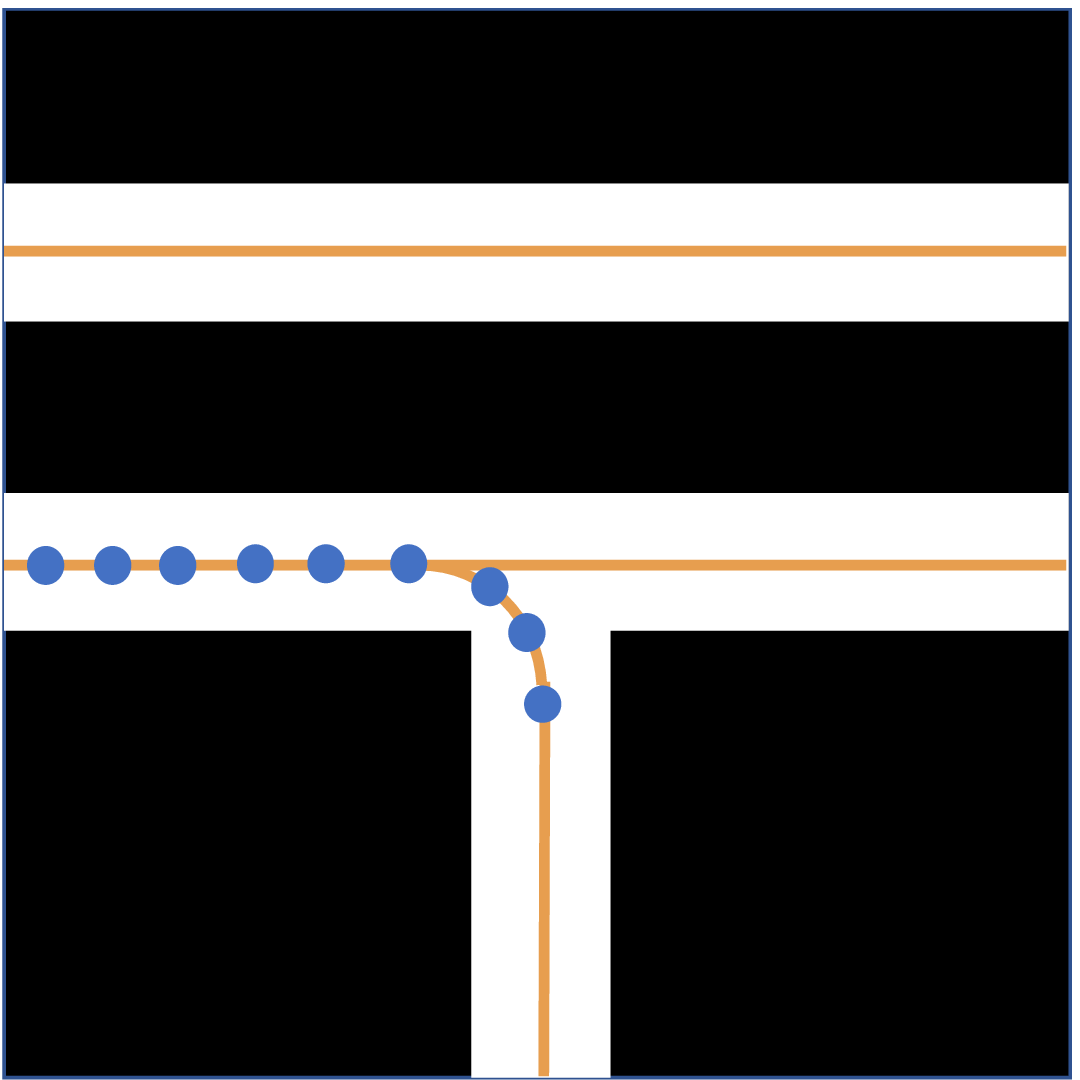}

}
\par\end{raggedleft}
\caption{(a) Right turn (b) Same right turn rotated by 90°\label{fig:trajectories}}
\end{figure}

The benefit of using the vehicle coordinate system is that it reduces
complexity of the input data. In the vehicle coordinate system, the
two trajectories depicted above are identical, whereas in global coordinates
they are not.

The ground truth for time step $t$ is $\boldsymbol{x}^{\left\langle t+\Delta\right\rangle }$,
i.e. the position of the ego vehicle $\text{\ensuremath{\Delta}}$
time steps ahead. The ground truth position is expressed in the vehicle
coordinate system of the current time step $t$. Since we are in the
unsupervised learning setting, these points do not come with any labels.
For each time step $t$ we can hence define the following input output
pair: $(\mathscr{\mathscr{T}}^{\left\langle 0:t\right\rangle },\ \mathcal{\mathscr{M}}^{\left\langle t\right\rangle };\ \boldsymbol{x}^{\left\langle t+\text{\ensuremath{\Delta}}\right\rangle })$.

In this paper the ground truth is defined only as the future position
of the ego vehicle. It can, however, be extended to encompass further
information, i.e. predicting the future heading angle.

\subsection{Loss Function}

In the following, we will denote random variables by an uppercase
letter, while their realizations will be denoted by a lowercase letter.

We wish to model the data in a generative manner\textit{}\footnote{\textit{A generative model includes the distribution of the data itself,
and tells you how likely a given example is.}} by approximating the following conditional density function:
\begin{multline}
f\left(\boldsymbol{X}^{\left\langle t+\text{\ensuremath{\Delta}}\right\rangle }\ |\ \mathscr{T}^{\left\langle 0:t\right\rangle },\ \mathcal{\mathscr{M}}^{\left\langle t\right\rangle }\right)=\\
f\left(X^{\left\langle t+\text{\ensuremath{\Delta}}\right\rangle },\ Y^{\left\langle t+\text{\ensuremath{\Delta}}\right\rangle }\ |\ \mathscr{T}^{\left\langle 0:t\right\rangle },\ \mathcal{\mathscr{M}}^{\left\langle t\right\rangle }\right),
\end{multline}
which gives the probabilities of all possible positions at time step
$t+\text{\ensuremath{\Delta}}$, given the input trajectory $\mathscr{T}^{\left\langle 0:t\right\rangle }$
up to time step t and the static map $\mathscr{M}^{\left\langle t\right\rangle }$
at time step $t$. Please note that $X^{\left\langle t+\text{\ensuremath{\Delta}}\right\rangle }$,
$Y^{\left\langle t+\text{\ensuremath{\Delta}}\right\rangle }$ are
expressed in the vehicle coordinate system of time step $t$.

We model the data by specifying a Gaussian mixture model for each
time step $t$:
\begin{multline}
p\left(\boldsymbol{X}^{\left\langle t+\text{\ensuremath{\Delta}}\right\rangle }\ |\ \mathscr{T}^{\left\langle 0:t\right\rangle },\ \mathcal{\mathscr{M}}^{\left\langle t\right\rangle };\ \theta_{1},\ \theta_{2}\right)=\\
\sum_{j=1}^{k}p\left(\boldsymbol{X}^{\left\langle t+\text{\ensuremath{\Delta}}\right\rangle }\ |\ Z^{\left\langle t+\text{\ensuremath{\Delta}}\right\rangle }=j,\ \mathcal{\mathscr{T}}^{\left\langle 0:t\right\rangle },\ \mathcal{\mathscr{M}}^{\left\langle t\right\rangle };\ \theta_{1}\right)\\
\cdot p\left(Z^{\left\langle t+\text{\ensuremath{\Delta}}\right\rangle }=j\ |\ \mathscr{T}^{\left\langle 0:t\right\rangle },\ \mathcal{\mathscr{M}}^{\left\langle t\right\rangle };\ \theta_{2}\right),\hphantom{dddddd}
\end{multline}
where $Z^{\left\langle t+\text{\ensuremath{\Delta}}\right\rangle }$
is a discrete latent random variable, which can take on $k$ different
values. The $\theta_{1},\ \theta_{2}$ denote the parameters of the
different parts of the Gaussian mixture model. The probability distribution:
\[
p(Z^{\left\langle t+\text{\ensuremath{\Delta}}\right\rangle }=j\ |\ \mathscr{T}^{\left\langle 0:t\right\rangle },\ \mathcal{\mathscr{M}}^{\left\langle t\right\rangle };\ \theta_{2})
\]
in the upper equation is characterized by $k$ probability values
$\phi_{j}$ satisfying $\phi_{j}\ge0$ and $\sum_{i=1}^{k}\phi_{j}=1$.
Furthermore for a Gaussian mixture model we assume: 
\[
\boldsymbol{X}^{\left\langle t+\text{\ensuremath{\Delta}}\right\rangle }\ |\ Z^{\left\langle t+\text{\ensuremath{\Delta}}\right\rangle }=j\ \sim\ \mathcal{N}(\boldsymbol{\mu}_{j},\ \boldsymbol{\Sigma}_{j}).
\]
In the subsequent formulas, for the sake of brevity, we won\textquoteright t
denote explicitly the dependency on $\mathscr{T}^{\left\langle 0:t\right\rangle }$,
$\mathcal{\mathscr{M}}^{\left\langle t\right\rangle }$ when there
is no risk of ambiguity.

Assuming that the $m$ training examples were generated independently,
we can write down the log-likelihood of the parameters $\theta_{1}$
and $\theta_{2}$ for a single time step $t$ as:
\begin{multline}
\mathcal{\ensuremath{\ell}}\left(t,\theta_{1},\theta_{2}\right)=\sum_{i=1}^{m}\log p\left(\boldsymbol{x}^{\left\langle t+\text{\ensuremath{\Delta}}\right\rangle \left(i\right)}\ |\ \ldots\ ;\ \theta_{1},\theta_{2}\right)\\
=\sum_{i=1}^{m}\log\sum_{j=1}^{k}\left[p\left(\boldsymbol{x}^{\left\langle t+\text{\ensuremath{\Delta}}\right\rangle \left(i\right)}\ |\ Z^{\left\langle t+\text{\ensuremath{\Delta}}\right\rangle }=j,\ \ldots\ ;\ \theta_{1}\right)\right.\\
\left.\cdot p\left(Z^{\left\langle t+\text{\ensuremath{\Delta}}\right\rangle }=j\ |\ \ldots\ ;\ \theta_{2}\right)\right],\hphantom{ddddd}
\end{multline}
where the superscripts in round brackets denote the sample number.
We will design the latent random variable $Z^{\left\langle t+\text{\ensuremath{\Delta}}\right\rangle }$
in such a way that its true values $z^{\left\langle t+\text{\ensuremath{\Delta}}\right\rangle \left(i\right)}$
are known beforehand and can easily be deduced from the ground truth
$\boldsymbol{x}^{\left\langle t+\text{\ensuremath{\Delta}}\right\rangle \left(i\right)}$
. We can then write the log-likelihood as:
\begin{multline}
\mathcal{\ensuremath{\ell}}\left(t,\theta_{1},\theta_{2}\right)=\sum_{i=1}^{m}\left[\log p\left(\boldsymbol{x}^{\left\langle t+\text{\ensuremath{\Delta}}\right\rangle \left(i\right)}\ |\ z^{\left\langle t+\text{\ensuremath{\Delta}}\right\rangle \left(i\right)},\ \ldots\ ;\theta_{1}\right)\right.\\
\left.+\log p\left(z^{\left\langle t+\text{\ensuremath{\Delta}}\right\rangle \left(i\right)}\ |\ \ldots\ ;\theta_{2}\right)\right],\hphantom{dddd}
\end{multline}
or using the ground truth of $Z^{\left\langle t+\text{\ensuremath{\Delta}}\right\rangle }$
we can rewrite the above expression as:
\begin{align}
\mathcal{\ensuremath{\ell}}\left(t,\theta_{1},\theta_{2}\right)=\sum_{i=1}^{m}\sum_{j=1}^{k}1\left\{ {\scriptstyle z^{\left\langle t+\text{\ensuremath{\Delta}}\right\rangle \left(i\right)}=j}\right\} \cdot\log p\left(\vphantom{\boldsymbol{x}^{\left\langle t+\text{\ensuremath{\Delta}}\right\rangle \left(i\right)}}j\ |\ \ldots\ ;\theta_{2}\right)\nonumber \\
+\sum_{i=1}^{m}\sum_{j=1}^{k}1\left\{ {\scriptstyle z^{\left\langle t+\text{\ensuremath{\Delta}}\right\rangle \left(i\right)}=j}\right\} \cdot\log p\left(\boldsymbol{x}^{\left\langle t+\text{\ensuremath{\Delta}}\right\rangle \left(i\right)}\ |\ j,\ \ldots\ ;\theta_{1}\right),
\end{align}
where $1\{\ldots\}$ is the indicator function indicating from which
Gaussian each sample had come. Plugging in the definition of a multivariate
Gaussian distribution and assuming $X^{\left\langle t+\text{\ensuremath{\Delta}}\right\rangle }$
and $Y^{\left\langle t+\text{\ensuremath{\Delta}}\right\rangle }$
to be independent we get:
\begin{align}
\mathcal{\ensuremath{\ell}}\left(t,\theta_{1},\theta_{2}\right)=\sum_{i=1}^{m}\sum_{j=1}^{k}1\left\{ {\scriptstyle z^{\left\langle t+\text{\ensuremath{\Delta}}\right\rangle \left(i\right)}=j}\right\} \cdot\log p\left(j\ |\ \ldots\ ;\theta_{2}\right)\nonumber \\
-\sum_{i=1}^{m}\sum_{j=1}^{k}{\scriptstyle \left[\frac{\left(x^{\left\langle t+\text{\ensuremath{\Delta}}\right\rangle \left(i\right)}-\mu_{x,j}^{\left\langle t+\text{\ensuremath{\Delta}}\right\rangle \left(i\right)}\right)^{2}}{2\cdot\left(\sigma_{x,j}^{\left\langle t+\text{\ensuremath{\Delta}}\right\rangle \left(i\right)}\right)^{2}}+\frac{\left(y^{\left\langle t+\text{\ensuremath{\Delta}}\right\rangle \left(i\right)}-\mu_{y,j}^{\left\langle t+\text{\ensuremath{\Delta}}\right\rangle \left(i\right)}\right)^{2}}{2\cdot\left(\sigma_{y,j}^{\left\langle t+\text{\ensuremath{\Delta}}\right\rangle \left(i\right)}\right)^{2}}\hphantom{dd}\right.}\label{eq:prognet_7}\\
\left.+\log\sigma_{x,j}^{\left\langle t+\text{\ensuremath{\Delta}}\right\rangle \left(i\right)}+\log\sigma_{y,j}^{\left\langle t+\text{\ensuremath{\Delta}}\right\rangle \left(i\right)}+\textrm{c}\vphantom{{\scriptstyle \left(\frac{x^{\left\langle t+\text{\ensuremath{\Delta}}\right\rangle \left(i\right)}}{\sigma_{x,j}^{\left\langle t+\text{\ensuremath{\Delta}}\right\rangle \left(i\right)}}\right)^{2}}}\ \right]\cdot1\left\{ {\scriptstyle z^{\left\langle t+\text{\ensuremath{\Delta}}\right\rangle \left(i\right)}=j}\right\} ,\nonumber 
\end{align}
with constant $\textrm{c}\textrm{=}\log2\pi$. The final cost function
is summed over all time steps $t$:
\[
\textrm{cost}=-\sum_{t=1}^{T_{\textrm{max}}}\mathcal{\ensuremath{\ell}}\left(t,\theta_{1},\theta_{2}\right),
\]
where $T_{\textrm{max}}$ is the number of time steps.

\subsection{Choice of Latent Variable}

\begin{figure}[tbh]
\begin{centering}
\includegraphics[viewport=0bp 25bp 367bp 340bp,scale=0.35]{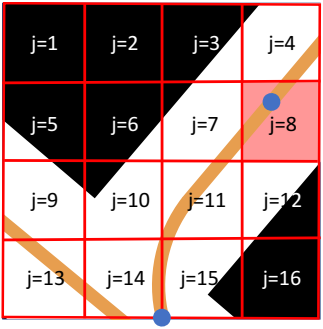}
\par\end{centering}
\caption{Subdivision of the static mask into $N\times N$ subregions,\newline$\protect\phantom{dddddd}$where
$N=4$\label{fig:subdivision}}
\end{figure}

We have to choose the latent variable $z^{\left\langle t+\text{\ensuremath{\Delta}}\right\rangle }$
in a way, such that its ground truth value can be obtained beforehand.
To this end we subdivide the static map into $N\times N$ subregions
and assign an index $j=1,\text{\dots},N^{2}$ to each of the subregions,
as depicted in Fig.\ \ref{fig:subdivision}.

We can now easily determine in which of the $N^{2}$ different subregions
the ground truth future position $\boldsymbol{x}^{\left\langle t+\text{\ensuremath{\Delta}}\right\rangle \left(i\right)}$
is located. Choosing the subregion $j$ as our latent variable hence
fulfills the required criteria.

\subsection{Interpretation}

Using the above definition of the latent variable let us dissect equation
(\ref{eq:prognet_7}) to gain more insight. We will start with the
first term:
\begin{equation}
-\sum_{i,j=1}^{m,k}1\left\{ {\scriptstyle z^{\left\langle t+\text{\ensuremath{\Delta}}\right\rangle \left(i\right)}=j}\right\} \cdot\log p\left(j\ |\ \mathscr{T}^{\left\langle 0:t\right\rangle \left(i\right)},\ \mathcal{\mathscr{M}}^{\left\langle t\right\rangle \left(i\right)};\theta_{2}\right),\label{eq:prognet_8}
\end{equation}
which is recognized as the cross-entropy loss in a multiclass setting
with $k=N^{2}$ classes. Hence our choice of the latent variable leads
to a cost function which contains a classification problem.

The cross-entropy loss is used in neural networks which have softmax
activations in the output layer. Hence we will realize $p(j\ |\ \mathcal{\mathscr{T}}^{\left\langle 0:t\right\rangle \left(i\right)},\ \mathcal{\mathscr{M}}^{\left\langle t\right\rangle \left(i\right)};\ \theta_{2})$
by a neural network with inputs $\mathcal{\mathscr{T}}^{\left\langle 0:t\right\rangle \left(i\right)}$
and $\mathscr{M}^{\left\langle t\right\rangle \left(i\right)}$ and
a softmax activation at the output layer. Please remember that this
part of the neural network outputs the probability distribution of
the discrete latent variable $Z^{\left\langle t+\text{\ensuremath{\Delta}}\right\rangle }$,
which is characterized by $k=N^{2}$ probability values $\phi_{j}$
satisfying $\phi_{j}\ge0$ and $\sum_{i=1}^{k}\phi_{j}=1$. In the
following we will hence denote the output of this network as $\phi_{j}^{\left\langle t+\text{\ensuremath{\Delta}}\right\rangle }$.

Each value $j$ which the latent variable $z^{\left\langle t+\text{\ensuremath{\Delta}}\right\rangle \left(i\right)}$
can assume, corresponds to a different spatial region. As has been
shown in section.\ \ref{sec:Introduction}, the future positions
of the ego vehicle are unequally distributed, see Fig.\ \ref{fig:imbalanced-dataset},
which makes (\ref{eq:prognet_8}) a highly imbalanced classification
problem. However, since we are in the classification domain, we have
powerful tools to combat this issue, e.g. focal loss \cite{Lin2018}.

Let us come back to the first part of the cost function:
\begin{multline}
\sum_{i,j=1}^{m,k}{\scriptstyle \left[\frac{\left(\mu_{x,j}^{\left\langle t+\text{\ensuremath{\Delta}}\right\rangle \left(i\right)}-x^{\left\langle t+\text{\ensuremath{\Delta}}\right\rangle \left(i\right)}\right)^{2}}{2\cdot\left(\sigma_{x,j}^{\left\langle t+\text{\ensuremath{\Delta}}\right\rangle \left(i\right)}\right)^{2}}+\frac{\left(\mu_{y,j}^{\left\langle t+\text{\ensuremath{\Delta}}\right\rangle \left(i\right)}-y^{\left\langle t+\text{\ensuremath{\Delta}}\right\rangle \left(i\right)}\right)^{2}}{2\cdot\left(\sigma_{y,j}^{\left\langle t+\text{\ensuremath{\Delta}}\right\rangle \left(i\right)}\right)^{2}}\right.}\\
{\scriptstyle \left.{\displaystyle +\log\sigma_{x,j}^{\left\langle t+\text{\ensuremath{\Delta}}\right\rangle \left(i\right)}+\log\sigma_{y,j}^{\left\langle t+\text{\ensuremath{\Delta}}\right\rangle \left(i\right)}}\right]}\cdot1\left\{ {\scriptstyle z^{\left\langle t+\text{\ensuremath{\Delta}}\right\rangle \left(i\right)}=j}\right\} .\label{eq:prognet_9}
\end{multline}
First of all please note that by design the ground truth value $\boldsymbol{x}^{\left\langle t+\text{\ensuremath{\Delta}}\right\rangle \left(i\right)}$
is contained in the subregion indicated by $z^{\left\langle t+\text{\ensuremath{\Delta}}\right\rangle \left(i\right)}$,
see Fig.\ \ref{fig:subdivision}. Because of the indicator function
in (\ref{eq:prognet_9}), for each sample $i$ only one Gaussian component
$\boldsymbol{\mu}_{j}^{\left\langle t+\text{\ensuremath{\Delta}}\right\rangle \left(i\right)}$,
$\boldsymbol{\sigma}_{j}^{\left\langle t+\text{\ensuremath{\Delta}}\right\rangle \left(i\right)}$
is active, with the same index as the selected subregion $j=z^{\left\langle t+\text{\ensuremath{\Delta}}\right\rangle \left(i\right)}$.
During training, the Gaussian components will therefore learn to correspond
to different subregions of the static map.

In order to minimize the above expression, the means $\boldsymbol{\mu}_{j}^{\left\langle t+\text{\ensuremath{\Delta}}\right\rangle \left(i\right)}$
of the Gaussian components must come as close as possible to $\boldsymbol{x}^{\left\langle t+\text{\ensuremath{\Delta}}\right\rangle \left(i\right)}$
, hence (\ref{eq:prognet_9}) can be interpreted as a (weighted) regression
problem with $k=N^{2}$ regressors covering different spatial positions.
The sigma terms $\boldsymbol{\sigma}_{j}^{\left\langle t+\text{\ensuremath{\Delta}}\right\rangle \left(i\right)}$
are capturing how much noise there is in the outputs \cite{Kendall2018}.
Since each regressor is covering different spatial positions, this
helps to combat the imbalanced data problem even further.
\begin{figure}[H]
\begin{centering}
\includegraphics[viewport=0bp 30bp 367bp 330bp,scale=0.35]{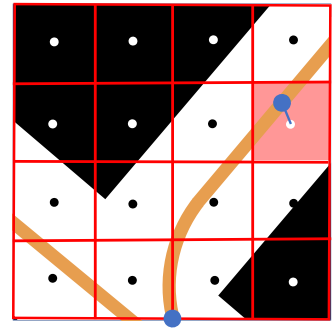}
\par\end{centering}
\caption{Grid defined by the center positions of each subregion,\newline$\protect\phantom{dddddd}$where
$N=4$\label{fig:subdivision-offset}}
\end{figure}

In the next step, we will simplify the regression problem, thus emphasizing
the classification task even further. Please note, that the classification
part (\ref{eq:prognet_8}) is already predicting the future position
of the ego vehicle, however quantized to the resolution of the subdivision.
We can hence design the regression part (\ref{eq:prognet_9}) only
to refine that prediction by providing an offset to the actual position.
To this end, we define a grid consisting of the center positions of
each subregion: $(\textrm{center}_{x,j},\textrm{center}_{y,j})$,
with $j=1,\ldots,N^{2}$. Now we can substitute in (\ref{eq:prognet_9}):
\begin{align}
\mu_{x,j}^{\left\langle t+\text{\ensuremath{\Delta}}\right\rangle \left(i\right)} & =\text{\ensuremath{\Delta}}\mu_{x,j}^{\left\langle t+\text{\ensuremath{\Delta}}\right\rangle \left(i\right)}+\textrm{center}_{x,j}\nonumber \\
\mu_{y,j}^{\left\langle t+\text{\ensuremath{\Delta}}\right\rangle \left(i\right)} & =\text{\ensuremath{\Delta}}\mu_{y,j}^{\left\langle t+\text{\ensuremath{\Delta}}\right\rangle \left(i\right)}+\textrm{center}_{y,j}\label{eq:prognet_10}
\end{align}
This further reduces the complexity of the regression problem, since
it has to predict only offsets, see Fig.\ \ref{fig:subdivision-offset}.

\subsection{Inference Time\label{subsec:Inference-Time}}

\begin{figure}[tbh]
\begin{centering}
\subfloat[]{\begin{centering}
\includegraphics[viewport=0bp 0bp 152bp 132.1739bp,scale=0.8]{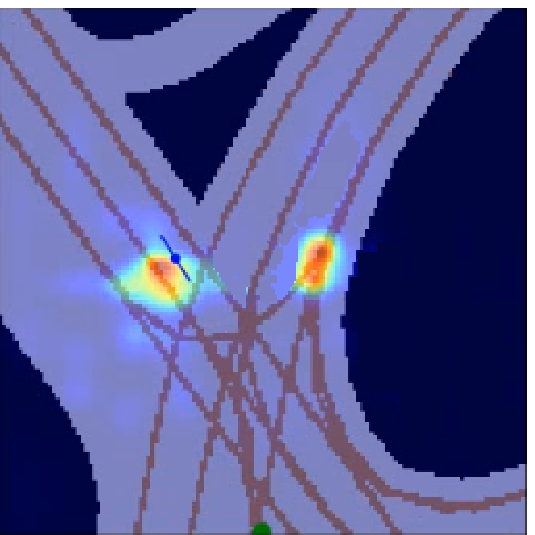}
\par\end{centering}
}\subfloat[]{\begin{centering}
\includegraphics[viewport=0bp 0bp 152bp 132.1739bp,scale=0.8]{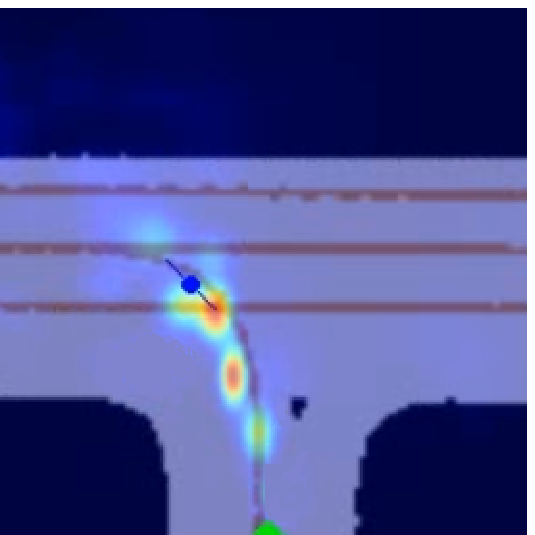}
\par\end{centering}
}
\par\end{centering}
\caption{Probability heatmap $p(\boldsymbol{X}^{\left\langle t+\text{\ensuremath{\Delta}}\right\rangle }\ |\ \mathscr{T}^{\left\langle 0:t\right\rangle },\ \mathcal{\mathscr{M}}^{\left\langle t\right\rangle };\ \theta_{1},\ \theta_{2})$\newline$\protect\phantom{ddddd}$
showing multimodality (a) and impact of static map (b)\label{fig:Multimodality}}
\end{figure}

At each time step $t$ the neural network outputs the parameters of
a Gaussian mixture model:
\[
\boldsymbol{\mu}_{j}^{\left\langle t+\text{\ensuremath{\Delta}}\right\rangle \left(i\right)},\ \boldsymbol{\sigma}_{j}^{\left\langle t+\text{\ensuremath{\Delta}}\right\rangle \left(i\right)}\ \mathrm{and}\ \phi_{j}^{\left\langle t+\text{\ensuremath{\Delta}}\right\rangle \left(i\right)}\ \mathrm{with}\ j=1,\text{\dots},N^{2}.
\]
This model estimates the probability of the vehicle of being at position
$\boldsymbol{X}^{\left\langle t+\text{\ensuremath{\Delta}}\right\rangle }$
in $\text{\ensuremath{\Delta}}$ time steps:
\begin{equation}
p\left(\boldsymbol{X}^{\left\langle t+\text{\ensuremath{\Delta}}\right\rangle }\ |\ \mathscr{T}^{\left\langle 0:t\right\rangle \left(i\right)},\ \mathcal{\mathscr{M}}^{\left\langle t\right\rangle \left(i\right)};\ \theta_{1},\ \theta_{2}\right),\label{eq:prognet_11}
\end{equation}
given the input trajectory $\mathscr{T}^{\left\langle 0:t\right\rangle }$
and the static map $\mathscr{M}^{\left\langle t\right\rangle }$.
We can visualize (\ref{eq:prognet_11}) by plugging values for $\boldsymbol{X}^{\left\langle t+\text{\ensuremath{\Delta}}\right\rangle }$
located on a dense uniform grid, see Fig.\ \ref{fig:Multimodality}.
The position of the vehicle at the current step is in the middle bottom
position (green point). The ground truth position and orientation
of the ego-vehicle in 2000 ms is shown as the blue arrow.

In many applications, however, instead of a heatmap we want to have
a list containing the most probable future positions. The list should
contain only distinct positions, thus reflecting the multimodality
of the future.

As can be seen in Fig.\ \ref{fig:Multimodality} (a), the $N^{2}$
predictions made by our approach tend to form distinct clusters. Out
of each cluster we want to keep only the most probable prediction
and get rid of all nearly identical ones. Non-maximum suppression
(NMS) is a way to make sure that each position is detected only once
\cite{Bodla2017}. To easily apply the standard NMS, we will interpret
the $\boldsymbol{\mu}_{j}^{\left\langle t+\text{\ensuremath{\Delta}}\right\rangle \left(i\right)}$
as the centers of bounding boxes and choose the bounding box sizes
to be proportional to $\boldsymbol{\sigma}_{j}^{\left\langle t+\text{\ensuremath{\Delta}}\right\rangle \left(i\right)}$
. The confidence required by NMS is obtained by using the fact that
by design the Gaussian components are mostly bound to their corresponding
subregions, yielding:
\[
\underset{\boldsymbol{X}^{\left\langle t+\text{\ensuremath{\Delta}}\right\rangle }\in\ \mathrm{subregion}{}_{j}}{\mathrm{max}}\ p(\boldsymbol{X}^{\left\langle t+\text{\ensuremath{\Delta}}\right\rangle }\ |\ \mathscr{T}^{\left\langle 0:t\right\rangle \left(i\right)},\ \mathcal{\mathscr{M}}^{\left\langle t\right\rangle \left(i\right)})\thickapprox\phi_{j}^{\left\langle t+\text{\ensuremath{\Delta}}\right\rangle \left(i\right)}
\]
which gives the probability that the ego vehicle will be in subregion
$j$ in $\Delta$ time steps. Please note that the number of predictions
returned by NMS is changing, however at least one prediction is returned.
We denote the distinct predictions as $\boldsymbol{\mu}_{k,\mathrm{nms}}^{\left\langle t+\text{\ensuremath{\Delta}}\right\rangle \left(i\right)}$,
$k=1,\ldots,\mathrm{max}$, where $k=1$ is the most probable prediction,
$k=2$ is the second most probable prediction and so on.

\subsection{Network Architecture}

In this section we describe a basic architecture needed for realizing
the model described in the previous sections. As described in section
\ref{sec:Input-Output}, the input of the network consists both of
the sequence describing the dynamics of the vehicle $\{(\text{\ensuremath{\Delta}}x^{\left\langle t\right\rangle },\ \text{\ensuremath{\Delta}}y^{\left\langle t\right\rangle },\ v^{\left\langle t\right\rangle },\ h^{\left\langle t\right\rangle })\}_{t=0}^{T_{\mathrm{max}}}$
and the sequence of static maps $\{\mathcal{\mathscr{M}}^{\left\langle t\right\rangle }\}_{t=0}^{T_{\mathrm{max}}}$.

\begin{figure}[tbh]
\begin{centering}
\includegraphics[viewport=0bp 30bp 746bp 417bp,scale=0.33]{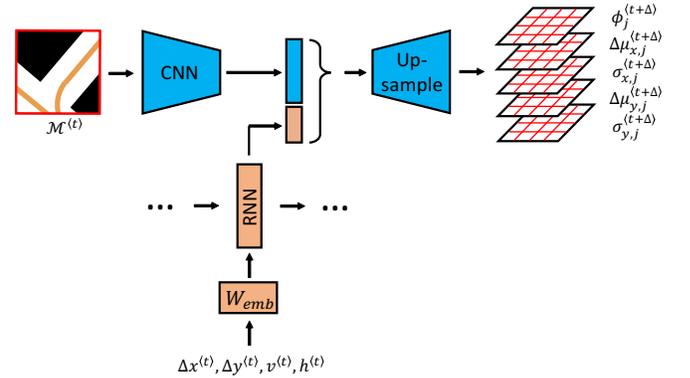}
\par\end{centering}
\caption{Basic architecture\label{fig:Basic-Architecture}}
\end{figure}

In order to generate sensible predictions, we need to aggregate the
past values of $\text{\ensuremath{\Delta}}x^{\left\langle t\right\rangle },\ \text{\ensuremath{\Delta}}y^{\left\langle t\right\rangle },\ v^{\left\langle t\right\rangle },\ h^{\left\langle t\right\rangle }$.
Hence a natural choice for processing of the input data (describing
the dynamics of the vehicle) is a recurrent neural network (RNN),
as depicted in Fig.\ \ref{fig:Basic-Architecture}. In the actual
implementation, a member of the broad family of RNN architectures
e.g. Long Short Term Memory (LSTM) and Gated Recurrent Unit (GRU),
as well as their variants, can be used. The low dimensional input
vector can optionally be embedded into a higher dimensional space.

The static map is represented as spatial data and is condensed to
a feature vector (representing semantical information of the static
map) using a convolutional neural network (CNN). The static maps only
act as constraints for possible predictions and hence do not require
an aggregation of past values, see Fig.\ \ref{fig:Basic-Architecture}.
To account for shifts and rotations of the static map arising from
the use of VCS, a spatial transformer network can be utilized \cite{Jaderberg2015}.

Both the output of the RNN and CNN are fed into the predictor, whose
output are five spatial maps consisting of $N$x$N$ cells. Each spatial
map represents a different component of the Gaussian mixture model:
$\phi_{j}^{\left\langle t+\Delta\right\rangle }$, $\Delta\mu_{x,j}^{\left\langle t+\Delta\right\rangle }$,
$\sigma_{x,j}^{\left\langle t+\text{\ensuremath{\Delta}}\right\rangle }$,
$\text{\ensuremath{\Delta}}\mu_{y,j}^{\left\langle t+\text{\ensuremath{\Delta}}\right\rangle }$
and $\sigma_{y,j}^{\left\langle t+\text{\ensuremath{\Delta}}\right\rangle }$.
In a subsequent postprocessing step we add a constant grid to the
coordinate offsets (delta terms), as described in (\ref{eq:prognet_10}).

\subsection*{Actual Implementation}

As mentioned before the context information we use, encompasses only
static context consisting of centerlines and the driveable area. For
this reason, only a simple architecture is required to achieve reasonable
results. More complex tasks e.g. with agents interaction, will require
more complicated architectures (feature pyramid networks, transposed
convolution etc.).

The input to the CNN network consists of a spatial map of size 128x128x2.
The centerlines and the driveable area are in separate feature maps.
The CNN network depicted in Fig.\ \ref{fig:Basic-Architecture} consists
of 5 layers. Each convolution layer is followed by a maximum pooling
layer to half the size of the feature map. We use ``tanh'' as activation
function in all layers. The output of the CNN is flattened to generate
a feature vector of size 256.

\begin{table}[tbh]
\begin{centering}
\subfloat[CNN]{\begin{centering}
\begin{tabular}{|c|c|c|}
\hline 
layer & size & nodes\tabularnewline
\hline 
\hline 
0 & 128x128 & 2\tabularnewline
\hline 
1 & 64x64 & 4\tabularnewline
\hline 
2 & 32x32 & 8\tabularnewline
\hline 
3 & 16x16 & 8\tabularnewline
\hline 
4 & 8x8 & 16\tabularnewline
\hline 
5 & 4x4 & 16\tabularnewline
\hline 
\end{tabular}
\par\end{centering}
}
\par\end{centering}
\begin{centering}
\subfloat[Embedding]{\begin{centering}
\begin{tabular}[b]{|c|c|}
\hline 
layer & nodes\tabularnewline
\hline 
\hline 
0 & 4\tabularnewline
\hline 
1 & 8\tabularnewline
\hline 
2 & 16\tabularnewline
\hline 
\end{tabular}
\par\end{centering}
}\subfloat[LSTM]{\begin{centering}
\begin{tabular}[b]{|c|c|}
\hline 
layer & nodes\tabularnewline
\hline 
\hline 
0 & 16\tabularnewline
\hline 
1 & 256\tabularnewline
\hline 
2 & 150\tabularnewline
\hline 
\end{tabular}
\par\end{centering}
}\subfloat[Upsampling]{\begin{centering}
\begin{tabular}[b]{|c|c|}
\hline 
layer & nodes\tabularnewline
\hline 
\hline 
0 & 256+150\tabularnewline
\hline 
1 & 256\tabularnewline
\hline 
2 & 128\tabularnewline
\hline 
3 & 500\tabularnewline
\hline 
\end{tabular}
\par\end{centering}
}
\par\end{centering}
\smallskip

\caption{Actual implementation\label{tab:CNN}}
\end{table}
The input to the recurrent part of the architecture consists of 4
values: the deltas between the current position and the last position,
velocity and the heading angle, as described in section.\ \ref{sec:Input-Output}.
Prior to feeding the input into the RNN, we use an embedding network
consisting of two layers. Again we use the ``tanh'' activation function.
The resulting embedding has a dimension of 16, see Tab.\ \ref{tab:CNN}
(b). The concrete implementation of the RNN network consists of a
LSTM with two layers, see Tab.\ \ref{tab:CNN} (c).

Finally, the output of the LSTM and CNN is concatenated to a vector
of size 256+150 which is fed to the Upsampling Network to obtain the
output maps. The upsamling layer is implemented as a stack of three
dense layers. The concrete implementation of the Upsampling Network
depicted in Fig.\ \ref{fig:Basic-Architecture} consists of 3 layers,
where the last layer is a linear layer. The output is reshaped to
form spatial maps of size 10x10x5.

\section{Evaluation on the Argoverse dataset}

\begin{figure}[tbh]
\begin{centering}
\subfloat[Right turn]{\begin{centering}
\includegraphics[viewport=0bp 0bp 201bp 209.8214bp,scale=0.55]{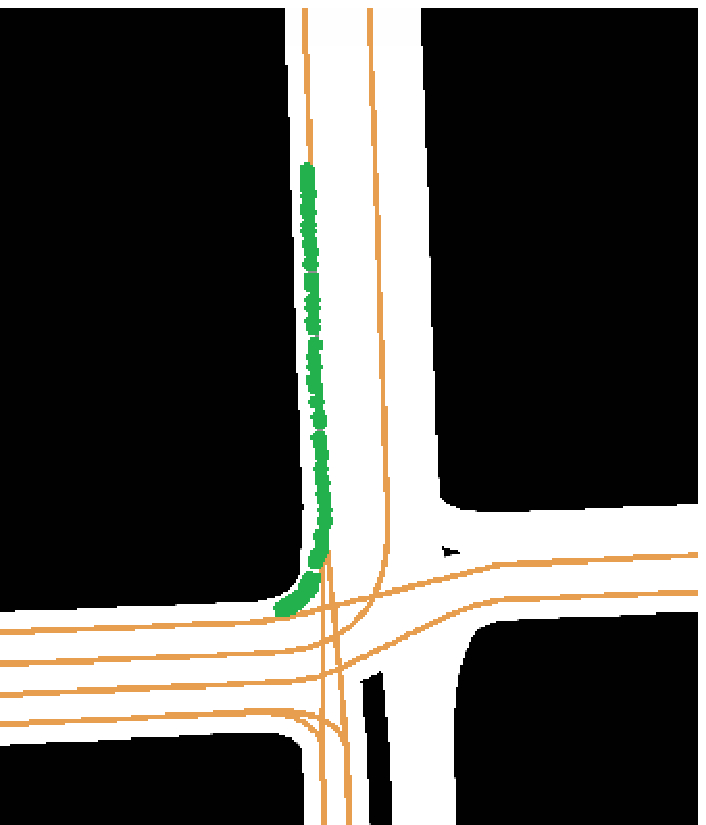}
\par\end{centering}
}\subfloat[Straight]{\begin{centering}
\includegraphics[viewport=0bp 0bp 201bp 209.8214bp,scale=0.55]{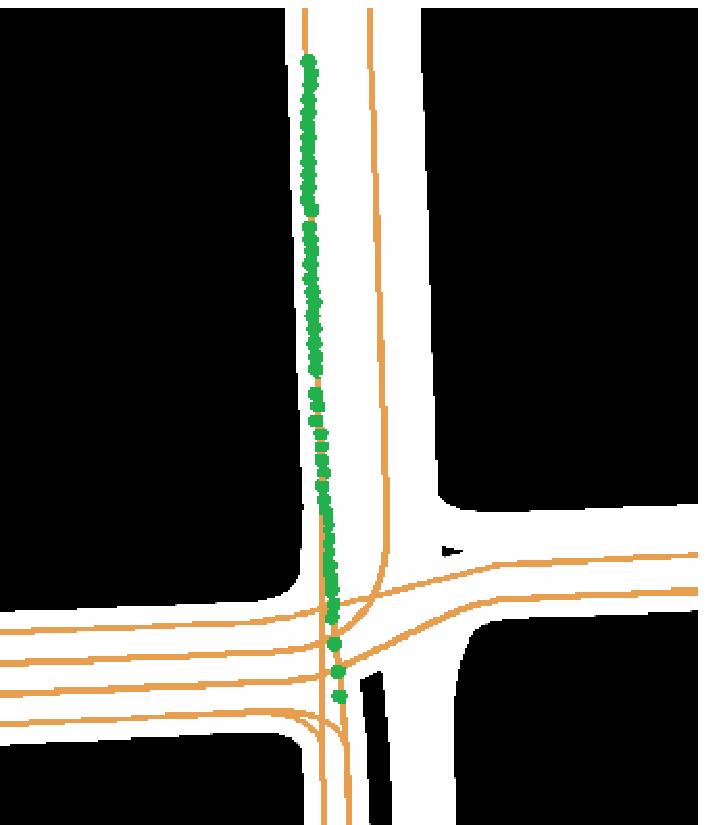}
\par\end{centering}
}
\par\end{centering}
\caption{Multimodality of Argoverse\label{fig:Multimodality-of-Argoverse}}
\end{figure}

In this section we will describe the evaluation of the presented approach
using the Argoverse dataset \cite{Chang2019}. Argoverse contains
a dataset for motion forecasting with 324,557 sequences and rich context
maps. Each sequence consists of exactly 50 samples varying in length
from 4 to 25 seconds. The motion forecasting dataset of Argoverse
was mined, in order to contain diverse scenarios e.g. managing an
intersection, slowing for a merging vehicle, accelerating after a
turn, stopping for a pedestrian on the road, etc. The number of tracks
in which a vehicle is traveling at nearly constant velocity (such
tracks are hardly a representation of real forecasting challenges)
is hence drastically diminished. Argoverse provides furthermore a
rich mutlimodality, as can be seen in Fig.\ \ref{fig:Multimodality-of-Argoverse}.

Our goal is to predict future positions of the ego-vehicle 2000ms
ahead given the static context. Since we operate on a sampling interval
of 100ms and we want to predict 2000ms into future, we choose the
ground truth position to lie $\text{\text{\ensuremath{\Delta}}}=20$
time steps ahead, see section \ref{sec:Input-Output}. Our overall
input trajectory hence consists of $T_{\mathrm{max}}=30$ time steps.
However not all motion forecasting sequences of Argoverse use a sampling
rate of 10Hz (sampling interval of 100ms). Out of the 324,557 sequences
we therefore filter 150,000 sequences with a length of approximately
5s, corresponding to a sampling rate of 10Hz. We split the data into
a training set consisting of 140,000 samples and a test set consisting
of 10,000 samples. We would like to stress that no balancing of the
data is performed whatsoever regarding the arrangement of the trajectories,
i.e. turns, straight lines etc.

We will evaluate our approach using the three most common metrics
used in motion forecasting: Average Displacement Error (ADE), Final
Displacement Error (FDE) and Minimum Average Displacement Error (minADE)
\cite{Rhinehart2018}.

The ADE metric is simply defined as the average euclidean distance
over all time steps and samples, calculated for the most probable
prediction $k=1$, which is obtained via non-maximum suppression (NMS)
as described in section \ref{subsec:Inference-Time}:
\begin{equation}
\mathrm{ADE}=\frac{1}{{\scriptstyle m\cdot\left(T_{\mathrm{max}}-4\right)}}\sum_{i=1}^{m}\sum_{t=5}^{T_{\mathrm{max}}}\left\Vert \boldsymbol{\mu}_{1,\mathrm{nms}}^{\left\langle t+\text{\ensuremath{\Delta}}\right\rangle \left(i\right)}-\boldsymbol{x}^{\left\langle t+\text{\ensuremath{\Delta}}\right\rangle \left(i\right)}\right\Vert ,
\end{equation}
where the length of trajectories is $T_{\mathrm{max}}=30$. Since
we are using an LSTM, we will ignore the first 5 time steps to account
for the warm-up phase.

As can be seen in Fig.\ \ref{fig:Multimodality}, in multimodal scenarios
the most likely prediction does not necessary need to be the correct
one. For this reason, the minADE metric was defined, which accounts
for multimodality:
\begin{align}
\mathrm{minADE}=\hphantom{\left\Vert \boldsymbol{\mu}_{k,\mathrm{nms}}^{\left\langle t+\text{\ensuremath{\Delta}}\right\rangle \left(i\right)}-\boldsymbol{x}^{\left\langle t+\text{\ensuremath{\Delta}}\right\rangle \left(i\right)}\right\Vert }\\
\frac{1}{{\scriptstyle m\cdot\left(T_{\mathrm{max}}-4\right)}}\sum_{i=1}^{m}\sum_{t=5}^{T_{\mathrm{max}}}\mathrm{\underset{k=1,2,3}{min}}\ \ \left\Vert \boldsymbol{\mu}_{k,\mathrm{nms}}^{\left\langle t+\text{\ensuremath{\Delta}}\right\rangle \left(i\right)}-\boldsymbol{x}^{\left\langle t+\text{\ensuremath{\Delta}}\right\rangle \left(i\right)}\right\Vert .\nonumber 
\end{align}
Here we choose the best prediction, w.r.t. to the euclidean distance,
out of a predefined number $K$ of predictions. Since our approach
only predicts single future positions and not whole trajectories,
we will limit the selection to $K=3$. Due to the NMS, the 3 predicted
positions should be quite distinct, i.e. not lying close to each other.
This way we prevent the scenario of achieving a good minADE just by
choosing the best prediction from a multitude of almost identical
predictions.

Finally, the last metric is the FDE:
\begin{eqnarray}
\mathrm{\mathrm{FDE}}=\frac{1}{{\scriptstyle m\cdot\left(T_{\mathrm{max}}-4\right)}}\sum_{i=1}^{m}\left\Vert \boldsymbol{\mu}_{k,\mathrm{nms}}^{\left\langle 5+\text{\ensuremath{\Delta}}\right\rangle \left(i\right)}-\boldsymbol{x}^{\left\langle 5+\text{\ensuremath{\Delta}}\right\rangle \left(i\right)}\right\Vert \\
\hphantom{dddddddd}+\left\Vert \boldsymbol{\mu}_{k,\mathrm{nms}}^{\left\langle T_{\mathrm{max}}+\text{\ensuremath{\Delta}}\right\rangle \left(i\right)}-\boldsymbol{x}^{\left\langle T_{\mathrm{max}}+\text{\ensuremath{\Delta}}\right\rangle \left(i\right)}\right\Vert .\nonumber 
\end{eqnarray}

\subsubsection*{Resampling of the original data}

Please note that all the metrics are evaluated at discrete time steps
$t$. However, real measurements are subject to jitter, the deviation
from the precise sample timing intervals of 100ms. Jitter, when unaccounted
for, may lead to a much higher prediction error than in reality. Imagine
the vehicle traveling at a constant velocity of 20 m/s (72 km/h).
A jitter of just $\pm$10ms leads to a covered distance of $\pm$0.2
meters. However, our architecture will learn to make predictions for
the average sampling rate of 100ms. We solve this issue by resampling
the original data to lie on a precise sampling grid of 100ms. However
we have to be cautious with the way in which we perform the resampling,
e.g. fitting a lower order polynome would smooth the trajectories,
thus making the prediction problem much easier. For this reason we
only use linear interpolation between two adjacent samples. This way
we do not suppress the measurement noise and do not introduce an advantage.
We utilize the resampled data both for training and evaluation.

\subsubsection*{Measurement Noise of the Argoverse Dataset}

So far we have neglected the fact that the ground truth values $\boldsymbol{x}^{\left\langle t\right\rangle \left(i\right)}$
contain statistical noise and other inaccuracies. At time $t$, a
measurement (or observation) $\boldsymbol{x}^{\left\langle t\right\rangle \left(i\right)}$
of the true vehicle position $\boldsymbol{\varkappa}^{\left\langle t\right\rangle \left(i\right)}$
is made according to: $\boldsymbol{x}^{\left\langle t\right\rangle \left(i\right)}=\boldsymbol{\varkappa}^{\left\langle t\right\rangle \left(i\right)}+\boldsymbol{v}^{\left\langle t\right\rangle \left(i\right)}$,
where $\boldsymbol{v}^{\left\langle t\right\rangle \left(i\right)}$
is the measurement noise which is assumed to be zero mean Gaussian
white noise (time steps are basically uncorrelated).

In order to correctly assess the prediction performance of our approach,
we have to determine the measurement noise $\sigma_{v}^{2}$ of the
Argoverse dataset. To approximate the true vehicle position $\boldsymbol{\varkappa}^{\left\langle t\right\rangle \left(i\right)}$,
we fit a curve $\tilde{C}\left(t\right)=(\tilde{x}\left(t\right),\ \tilde{y}\left(t\right))$
to the 50 times steps of the trajectories, where $\tilde{x}\left(t\right)$
and $\tilde{y}\left(t\right)$ are sixth order polynomials. To take
the jittering of the sampling rate into account, we use the exact
exact time steps $t$ (in ms) provided by the Argoverse dataset and
not just multiples of 100ms. In order to cope with possible outliers
we utilize the RANSAC (Random Sampling Consensus) \cite{Fischler1981}.
Next we calculate the ADE metric between the fitted trajectory and
the original data and obtain an estimate of the measurement noise
of $\sigma_{v}=0.46$ meters. Please note, that this value can be
interpreted as the lower bound of ADE for the Argoverse dataset.

Let us denote the true (unknown) prediction error as: 
\[
\mathrm{\mathbf{e}}_{p}^{\left\langle t+\text{\ensuremath{\Delta}}\right\rangle \left(i\right)}=\boldsymbol{\mu}_{1,\mathrm{nms}}^{\left\langle t+\text{\ensuremath{\Delta}}\right\rangle \left(i\right)}-\boldsymbol{\varkappa}^{\left\langle t+\text{\ensuremath{\Delta}}\right\rangle \left(i\right)},
\]
where $\boldsymbol{\varkappa}^{\left\langle t\right\rangle \left(i\right)}$
is the unknown true vehicle position. Since $\boldsymbol{\mu}_{1,\mathrm{nms}}^{\left\langle t+\text{\ensuremath{\Delta}}\right\rangle \left(i\right)}$
is calculated using only time steps $0,\ldots,t$, we can safely assume
$\mathrm{\mathbf{e}}_{p}^{\left\langle t+\text{\ensuremath{\Delta}}\right\rangle \left(i\right)}$
and $\boldsymbol{v}^{\left\langle t+\text{\ensuremath{\Delta}}\right\rangle \left(i\right)}$
to be statistically independent\footnote{The measurement noise of values lying 2000ms apart shall be independent.},
hence $\mathrm{ADE}=\sqrt{\sigma_{\mathrm{\mathbf{e}}_{p}}^{2}+\sigma_{v}^{2}}$.
In the following we will use this formula to estimate the metrics,
obtained if the true vehicle position $\boldsymbol{\varkappa}^{\left\langle t\right\rangle \left(i\right)}$
was accessible.

\begin{table}[tbh]
\begin{centering}
\subfloat[Metrics evaluated on ground truth \newline$\protect\phantom{ddd}$containing
measurement noise]{\begin{centering}
\begin{tabular}{|c|c|c|}
\hline 
ADE & minADE (K=3) & FDE\tabularnewline
\hline 
\hline 
0.93m & 0.78m & 0.96m\tabularnewline
\hline 
\end{tabular}
\par\end{centering}
}
\par\end{centering}
\begin{centering}
\subfloat[Estimates of metrics evaluated on\newline$\protect\phantom{ddd}$ground
truth without measurement\newline$\protect\phantom{ddd}$noise using
$\sigma_{\mathrm{\mathbf{e}}_{p}}=\sqrt{\mathrm{ADE}^{2}-\sigma_{v}^{2}}$]{\begin{centering}
\begin{tabular}{|c|c|c|}
\hline 
ADE & minADE (K=3) & FDE\tabularnewline
\hline 
\hline 
0.81m & 0.63m & 0.84m\tabularnewline
\hline 
\end{tabular}
\par\end{centering}
}
\par\end{centering}
\bigskip

\caption{Evaluation on linearly resampled data for $\Delta=20$ \cite{Chandra2019}}
\end{table}

\section{Conclusions and Future Work$\protect\phantom{}$}

In this paper, we have introduced a novel way to predict future positions
of a vehicle taking static context information into account. Our main
focus lay on the capability to make multimodal predictions with probabilities
assigned to them. The second goal was to cope with the imbalance of
data. We achieved both goals by introducing a generative probabilistic
model based on a Gaussian mixture model. A smart choice of the latent
variable allowed for the reformulation of the problem into a combination
of a classification problem and a simplified regression problem.

The first benefit of this formulation arose from the fact that we
obtained a classification problem, which allowed to combat the imbalanced
data problem by utilizing focal loss. The second benefit arose from
the fact that the regression part was spatially distributed, allowing
each regressor to specialize on a different subregion.

Our current focus of research lies in the extension of the proposed
approach to predict whole trajectories instead of single points in
time. In parallel we are examining methods to extend the context information
to not only contain the static context like driveable area, centerlines
and traffic signs, but also to contain dynamic context, i.e. other
agents.

\bibliographystyle{plain}
\bibliography{PrognoseNet}

\end{document}